\documentclass{article} 
\usepackage{iclr2026_conference,times}
\usepackage{graphicx}
\usepackage{xcolor}
\usepackage{wrapfig}
\usepackage{comment}
\usepackage{booktabs}
\usepackage[table]{xcolor} 
\definecolor{navy}{rgb}{0.0,0.0,0} 
\usepackage{colortbl}  
\usepackage{threeparttable}
\usepackage{multirow}
\usepackage[T1]{fontenc}
\usepackage{graphicx}
\usepackage{wrapfig}
\usepackage{caption}

\usepackage{amsmath,amsfonts,bm}

\def\eqref#1{equation~\ref{#1}}

\def\1{\bm{1}}

\DeclareMathAlphabet{\mathsfit}{\encodingdefault}{\sfdefault}{m}{sl}
\SetMathAlphabet{\mathsfit}{bold}{\encodingdefault}{\sfdefault}{bx}{n}

\usepackage{hyperref}
\usepackage{url}

\usepackage{ifthen}
\usepackage{xcolor}

\newcommand{\compilehidecomments}{false}

\ifthenelse{ \equal{\compilehidecomments}{true} }{%
	\newcommand{\wei}[1]{}
	\newcommand{\lina}[1]{}
}{
	\newcommand{\wei}[1]{{\color{blue}  [\text{Wei:} #1]}}
	\newcommand{\lina}[1]{{\color{red!80!black} [\text{Lina:} #1]}}
}

\title{Planner and Executor : \\Collaboration  between Discrete Diffusion \\And Autoregressive Models in Reasoning}

\author{
\textbf{Lina Berrayana}$^{1\dagger \ddagger}$, \textbf{Ahmed Heakl}$^{2\dagger}$, \textbf{Muhammad Abdullah Sohail}$^{2\dagger}$, \\
\textbf{Thomas Hofmann}$^{3}$, \textbf{Salman Khan}$^{2}$, \textbf{Wei Chen}$^{4 \ddagger}$ 
\thanks{† denotes equal contribution.   ‡ Corresponding authors (weic@microsoft.com, lina.berrayana@epfl.ch).} \\
\texttt{$^{1}$EPFL, $^{2}$MBZUAI,  $^{3}$ETH Zürich, $^{4}$Microsoft Research Asia }
}

\iclrfinalcopy 
\begin{document}

\maketitle

\begin{abstract}Current autoregressive language models (ARMs) achieve high accuracy but require long token sequences, making them costly. 
Discrete diffusion language models (DDLMs) enable parallel and flexible generation within a fixed number of steps and have recently emerged for their strong performance in complex reasoning and long-term planning tasks.
We present a study exploring hybrid architectures that couple DDLMs with ARMs to assess whether their collaboration can yield complementary benefits.
We first examine collaboration in text space, where one model plans the reasoning process and another executes the final answer based on that plan.
We then extend this setup to latent-space communication, introducing a learned projector that maps DDLM latents into the ARM’s embedding space, potentially bypassing some of the text-generation limitations of diffusion models.
We find that shifting DDLM$\to$ARM communication from text space to latent space yields significant accuracy gains, for example increasing from 27.0\% to 54.0\% on DART-5 and from 0.0\% to 14.0\% on AIME24.
We also find that combining a DDLM planner with an ARM executor can provide substantial computational savings with little to no impact on accuracy. For example, the latent-space pipeline, using 64 tokens for planning and roughly 5 for execution, surpasses Qwen3.1-7B on DART-5 and AIME, despite Qwen using 44 times more tokens.
Overall, our study offers new insights into reasoning with DDLMs and highlights their potential in hybrid architectures.
\end{abstract}

\section{Introduction}

\begin{wrapfigure}{r}{0.50\textwidth} 
    \centering
    \vspace{-4.05em}
    \includegraphics[width=\linewidth]{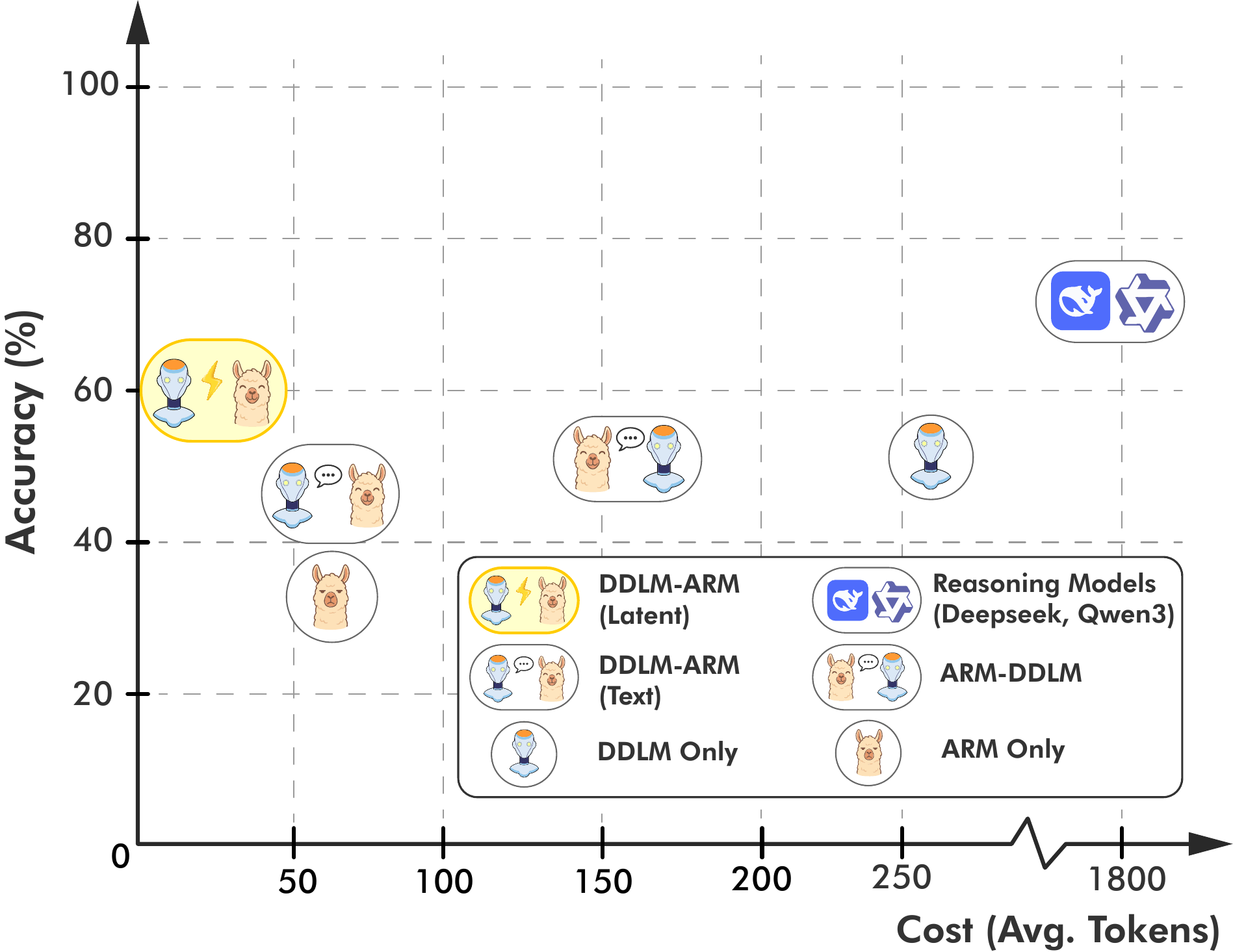}
    \caption{Accuracy–cost trade-offs across planner–executor configurations. 
    DDLM$\to$ARM, particularly with latent-space exchange, achieves higher reasoning accuracy 
    at lower token budgets compared to ARM-only, ARM$\to$DDLM, and reasoning models.}
    \label{fig:teaser}
\end{wrapfigure}

Over the past few years, autoregressive language models (ARMs) \citep{10.5555/944919.944966} have emerged as the dominant paradigm in natural language processing and artificial intelligence, achieving remarkable performance across a wide range of applications \citep[\emph{inter alia}]{openai2023gpt4,achiam2023gpt,anthropic2023claude,team2023gemini}. These models have achieved substantial progress in reasoning tasks through structured approaches such as step-by-step reasoning  \citep{2022arXiv220511916K,cobbe2021trainingverifierssolvemath}, chain-of-thought prompting \citep{wei2023chainofthoughtpromptingelicitsreasoning}, and plan-and-solve strategies \citep{2023arXiv230504091W}. More recently, specialized reasoning models \citep{xu2025largereasoningmodelssurvey}, such as DeepSeek-R1 \citep{guo2025deepseek} and Qwen3 \citep{yang2025qwen3}, have demonstrated state-of-the-art performance on challenging mathematical and logical benchmarks \citep{Wei2022ChainOT,sun2025challengingboundariesreasoningolympiadlevel}.

In parallel, discrete diffusion language models (DDLMs) \citep{yu2025discretediffusionlargelanguage,li2025surveydiffusionlanguagemodels} have recently attracted growing attention, spurred by findings that they can surpass ARMs in complex reasoning and planning tasks \citep{ye2025autoregressiondiscretediffusioncomplex}.  Novel approaches such as Diffusion-of-Thought \citep{ye2024diffusion}, hybrid strategies that integrate discrete and uniform diffusion \citep{vonrütte2025generalizedinterpolatingdiscretediffusion,li2025unifyingcontinuousdiscretetext}, and reinforcement learning for DDLMs \citep{zhao2025d1scalingreasoningdiffusion} highlight their potential in reasoning tasks. However, the quality of text generated by DDLMs, as measured by standard metrics such as perplexity, lags behind autoregressive approaches \citep{arriola2025block,gulrajani2023likelihood,sahoo2024simple}, limiting their applicability.

These two paradigms offer complementary strengths: autoregressive decoding produces fluent, coherent and human-comprehensible sequences through next-token prediction \citep{radford2019language,feng2025theoreticalbenefitlimitationdiffusion}, while diffusion enables flexible token generation, a property that recent studies suggest is particularly advantageous for planning tasks \citep{ye2023diffusion,zhang2023planner}, as well as offering potential benefits in self-correction \citep{hoogeboom2021argmax} and computational efficiency \citep{lou2023discrete}. What remains underexplored is how these families could \emph{collaborate} to solve reasoning tasks: Should planning be delegated to DDLMs while execution remains ARMs? Could information exchange in latent space between models enable collaborative improvements?

This paper is an \emph{investigation} of DDLM–ARM collaboration for reasoning. We study planner–executor pairings of these language models, and we compare two communication channels: (i) \textbf{text space}, where a planner emits a textual plan that conditions the executor, and (ii) \textbf{latent space}, where a learned projector maps diffusion representations directly to the executor’s embedding space. Figure~\ref{fig:teaser} illustrates the average accuracy–cost trade-offs observed across all experimental configurations, highlighting how latent-space collaboration shifts the efficiency frontier.

Our goal is \emph{not} to propose a single system that “solves long reasoning with fewer tokens.” Instead, we treat token budgets as a controlled variable and report token savings and their compute implications as an \emph{outcome} of the collaboration dynamics we uncover.

Our contributions are as follows:

\begin{enumerate}
    \item \textbf{Systematic study of planner–executor roles.} We evaluate four pairings (ARM$\rightarrow$ARM, ARM$\rightarrow$DDLM, DDLM$\rightarrow$ARM, DDLM$\rightarrow$DDLM) on diverse reasoning benchmarks, isolating when one model should plan and the other should execute.
    \item \textbf{Two collaboration channels.} We compare \emph{text-space} prompting against a \emph{latent-space} projector that maps diffusion states into the executor’s embedding space, quantifying when latent exchange yields superior performance.
    \item \textbf{Towards hybrid DDLM--ARM collaboration.} To our knowledge, we provide the first systematic study of \emph{hybrid} collaboration strategies between autoregressive and discrete diffusion language models \emph{focused on reasoning performance}, examining their impact through a ``plan-first, execute-after'' pipeline.
    
\end{enumerate}

Together, these results show when and how DDLMs and ARMs \emph{complement} each other for reasoning, and highlight design principles for future hybrid planners and executors, enabling explicit trade-offs between compute, fluency, and robustness instead of relying on longer token chains.






\section{Related Work}

\paragraph{Autoregressive reasoning.}
Autoregressive (AR) language models achieve strong reasoning accuracy, especially with chain-of-thought (CoT) prompting and its variants such as zero-shot CoT and self-consistency \citep{wei2022chainofthought,kojima2022zeroshot,wang2022selfconsistency}. These methods elicit multi-step, token-by-token traces that reliably boost accuracy on arithmetic, commonsense, and symbolic tasks. However, the sequential nature of AR decoding makes long CoT traces computationally expensive at inference time and can lead to verbosity or drift in long-horizon problems \citep{chowdhery2022palm}. This motivates exploring non-autoregressive mechanisms that decouple “thinking cost’’ from output length.

\paragraph{Discrete diffusion for language.}
Discrete diffusion language models generate text by iterative denoising with parallel token updates and flexible order, trading a fixed number of steps for global edits \citep{austin2021d3pm,li2022diffusionlm}. Early discrete diffusion models underperformed autoregressive (AR) models on standard language modeling metrics (e.g., perplexity) \citep{austin2021d3pm}, but subsequent work improved controllability and output quality through better objectives and guidance \citep{li2022diffusionlm}.
Conceptually, diffusion offers compute control (fixed steps) and global revision, yet practical challenges remain: weaker surface fluency and alignment compared to AR decoders, especially for long, well-formed rationales \citep{feng2025theoreticalbenefitlimitationdiffusion}. Recent diffusion-of-thought methods integrate CoT-style reasoning into the diffusion process and report competitive results on multi-step tasks, indicating growing viability of diffusion-based reasoning \citep{ye2024dot}.

\paragraph{Hybrid architectures.}  
A complementary line of research decouples \emph{planning} from \emph{execution} in modular agents: a high-level planner proposes a structured plan, and an executor implements it in the environment or produces the final answer \citep{erdogan2025planact}. Such factorization improves long-horizon reliability by allowing different modules to specialize \citep{wang2023planandsolvepromptingimprovingzeroshot}.  

Other studies have proposed approaches such as block diffusion \citep{arriola2025block}, which interpolate between ARMS and DDLMs, to address the two main challenges of discrete diffusion: generating sequences of arbitrary length and closing the perplexity gap with ARMs, while retaining the advantages of DDLMs. However, these works do not investigate the use of DDLMs as a planner for reasoning tasks. In our setting, we treat the fixed-length generation of discrete diffusion models not as a limitation but as a feature, exploring whether, under a constrained computational budget, DDLMs can collaborate with ARMs to improve reasoning performance efficiently.

\section{Methods}
\label{sec:methods}

We aim to investigate the potential benefits of fostering collaboration between ARMs and DDLMs 
through a planner–executor framework. 
We begin by defining the framework itself, before outlining the two collaboration strategies under consideration.

\paragraph{Planner–Executor Framework}

We use the terms \emph{planner} and \emph{executor} to describe two complementary roles:

\textbf{Planner.} We define a \emph{planner} as a language model whose output is intended to support the solution of a reasoning task, 
without producing the final answer.  
In a reasoning model, this corresponds to the ``thinking'' phase, if it were generated separately from the final response.  
For example, in step-by-step reasoning techniques such as \emph{Chain-of-Thought}~\citep{wei2023chainofthoughtpromptingelicitsreasoning}, 
the planner can be viewed as the component responsible for generating intermediate reasoning steps without the final prediction.  
More explicitly, in the \emph{Plan-and-Solve} prompting technique \citep{wang2023planandsolvepromptingimprovingzeroshot} which has been shown to improve performance on mathematical reasoning problems, the planner is the module that generates the plan.

 \textbf{Executor.} The \emph{executor} is a language model responsible for producing the final answer, given the original question and the planner’s output, without additional explicit reasoning.

\paragraph{Communication Channels}
We investigate two ways of transmitting information from the planner to the executor. 
These approaches differ in how they encode, transmit, and interpret the reasoning signal
:

 \textbf{Text-space collaboration. } 
In the first setup, the planner produces an explicit textual plan, which is then appended directly to the executor’s input prompt. This scheme is attractive for its simplicity: it requires no architectural changes, and it mirrors the widely adopted chain-of-thought paradigm, except that plan generation is handled by one model (e.g., DDLM), while a separate model (e.g., ARM) is responsible for finalizing the reasoning into a concrete answer. Another key advantage of text-based collaboration is its full interpretability, as intermediate reasoning steps remain visible and verifiable. However, one may question whether the effectiveness of this approach relies too strongly on the planner’s ability to generate fluent text. Diffusion-based models, in particular, may produce less coherent outputs, potentially degrading the quality of the prompt passed to the executor when acting as planners \citep{arriola2025block,gulrajani2023likelihood,sahoo2024simple}. This can result in higher sequence error rates, as reported by \cite{feng2025theoreticalbenefitlimitationdiffusion}, which may ultimately constrain their effectiveness in planning roles. While we do not claim that this will automatically degrade performance, we were inspired by the Chain of Continuous Thought approach \citep{hao2024traininglargelanguagemodels}, which suggests that reasoning and collaboration can occur in latent space rather than being confined to surface text. The idea that language models can collaborate without being constrained by linguistic fluency motivated the design of the following collaboration pathway.

\begin{figure}[t] 
    \centering 
    \includegraphics[width=0.9\textwidth]{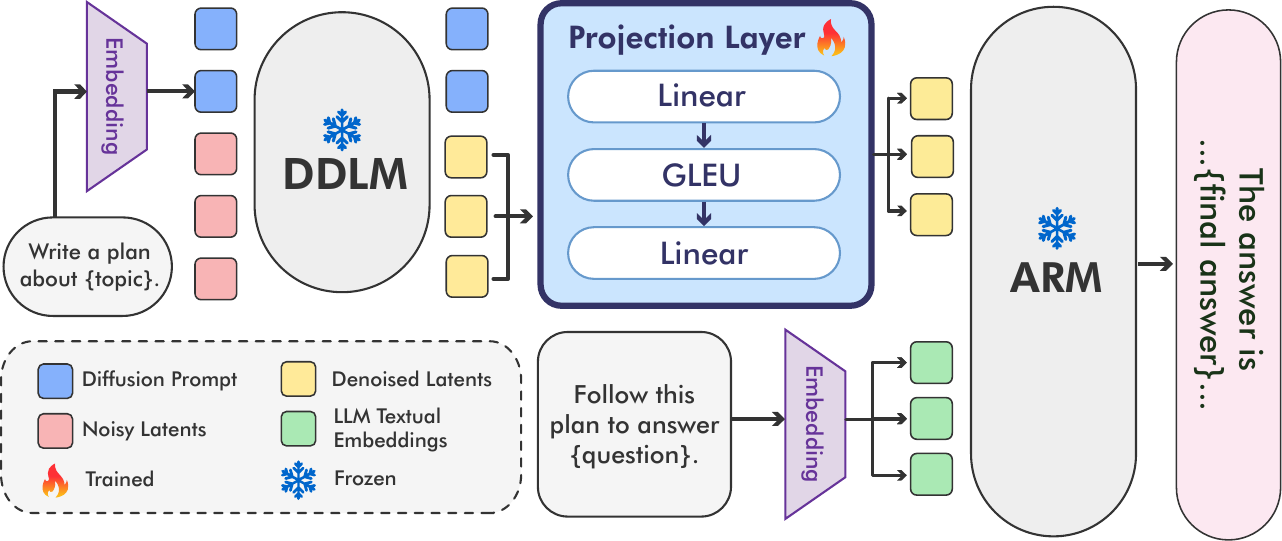} 
    \caption{Overview of the latent-space collaboration pipeline. A discrete diffusion language model (DDLM) generates a structured plan from noisy latents. The plan is projected directly into the autoregressive model (ARM) embedding space through a learned projection layer (latent space). The ARM then conditions on the plan and the question to produce the final answer. 
    } 
    \label{fig:projector} 
\end{figure}

\textbf{Latent-space collaboration.}
Figure~\ref{fig:projector} shows the latent-space configuration, where a DDLM-based planner generates plans directly in the latent space, and an ARM-based executor produces the final answers. Communication between the models is enabled via a learned projection layer that maps DDLM states into the ARM executor’s embedding space.
The projector consists of a Linear–GELU–Linear stack trained to align DDLM plans with LLM embeddings. During inference, the DDLM latents are projected directly into the executor’s hidden space, bypassing a raw textual plan.
While this approach sacrifices interpretability, since the intermediate reasoning steps are no longer visible, it allows for richer and more expressive communication between models. In fact, recent work has shown that diffusion models can encode the correct answer in their latent representations before surface-level decoding \citep{li2025diffusionlanguagemodelsknow}, suggesting that latent exchange may unlock deeper reasoning signals that text-based communication cannot capture.

\paragraph{End-to-End Pipeline}
The overall pipeline is given a reasoning question to answer. In the DDLM$\rightarrow$ARM configuration, the question is first processed by the DDLM, which iteratively denoises latent variables into a structured plan representation. This plan can then be communicated to the executor through two channels: as surface text appended to the question (\emph{text space}) or as a latent embedding projected directly into the executor’s representation space (\emph{latent space}). The executor, instantiated as an ARM, conditions on both the original question and the provided plan to produce the final answer.


\section{Collaboration Through Text Space}

\subsection{Experimental Setup}
\label{exp_setup}

This section details the experimental setup for the text-space collaboration presented in Section~\ref{sec:methods}.
To rigorously assess experimentally the benefits of collaboration between ARMs and DDLMs in this setting, we evaluate all the following possible combinations:

\textbf{ARM-only.} 
We first evaluate the performance of an ARM used in isolation, where it is directly prompted to answer the given questions without an explicit planning step.  

\textbf{ARM $\rightarrow$ ARM.}  
Next, we evaluate a setup where one ARM acts as the planner, generating intermediate reasoning or guidance, and another instance of the same model serves as the executor, producing the final answer.  

\textbf{DDLM-only and DDLM $\rightarrow$ DDLM.}  
In parallel, we first assess a DDLM in isolation as an executor-only baseline.  
We then evaluate a DDLM $\rightarrow$ DDLM setup, where two instances of the same DDLM are used, the first serving as planner and the second as executor.  

\textbf{ARM $\rightarrow$ DDLM and DDLM $\rightarrow$ ARM.}  
Finally, we examine collaborative setups between ARMs and DDLMs, alternating their roles.

\subsection{Models and Benchmarks}

\textbf{DDLMs.}  
We evaluate the setups described in Section \ref{exp_setup} using two recently released masked diffusion models (MDMs) from 2025, LLada-8B-Instruct \citep{nie2025largelanguagediffusionmodels} and Dream-v0-Instruct-7B \citep{ye2025dream7bdiffusionlarge}. We set the default sequence length to 256 tokens, providing the planner with enough capacity to generate reasoning plans while minimizing repetitions errors (see details in Appendix~\ref{sec:planner_repetition}).

\textbf{ARMs: Non-Reasoning Models.}
Alongside the DDLMs, we consider two autoregressive models (ARMs) of comparable size, Qwen2.5-7B-Instruct\citep{qwen25} and Llama-3.1-8B-Instruct\citep{touvron2023llamaopenefficientfoundation, dubey2024llama}, for a fair comparison. We also include two smaller ARMs, Llama-3.2-3B-Instruct\citep{touvron2023llamaopenefficientfoundation, dubey2024llama} and Qwen2.5-3B-Instruct\citep{qwen25}, to illustrate the results with smaller models.

\textbf{ARMs: Reasoning Models.} 
We additionally compare the previously described collaboration setups against reasoning models to contextualize their performance relative to state-of-the-art systems. 
Specifically, we evaluate two reasoning-oriented models: Qwen3-1.7B~\citep{qwen3} and DeepSeek-R1-Distill-Qwen-7B~\citep{qwen25}. 
The latter is a distilled variant trained to compress the reasoning ability of the original DeepSeek-R1 into a smaller Qwen-7B backbone. 
For simplicity, we refer to this distilled variant as the \emph{DeepSeek model} throughout the remainder of the paper, even though it is not the full original model.
Our objective in these experiments is not to match the performance of specialized reasoning models, but to study the potential of hybrid ARM-DDLM collaborations 
among general language models
when compared to strong state-of-the-art reasoning baselines.

\textbf{Benchmarks.} We evaluate on a diverse suite of reasoning benchmarks: ARC-E and ARC-C \citep{Clark2018ThinkYH}, science exam questions at Easy and Challenge difficulty; MMLU \citep{hendryckstest2021,hendrycks2021ethics}, spanning mathematics, history, computer science, and law; and AIME 2024 \citep{aime_1983_2024,aime_2024}, a high-school mathematics competition. We further include DART-1 through DART-5 \citep{tong2024dartmathdifficultyawarerejectiontuning}, a large-scale mathematical reasoning benchmark covering five difficulty levels. These benchmarks are standard in the literature and together provide broad coverage of reasoning domains. For each evaluation, we use the greater of 200 samples or the full benchmark size.


\subsection{Results}


Overall, the text-space experiments summarized in Table~\ref{tab:consolidated_performance} reveal several notable trends. 
\textbf{(i)} In general, \textsc{DDLM}-only configurations outperform their hybrid counterparts that combine \textsc{DDLM} and \textsc{ARM}, with a few exceptions such as 
LLada-8B~+~Qwen3.1-7B, where the \textsc{DDLM}$\to$\textsc{ARM} setup achieves slightly higher scores. This deviation is likely attributed to the strong standalone performance of the Qwen3.1-7B reasoning model itself, which surpasses its paired version when operating independently.  
\textbf{(ii)} Interestingly, the reverse configuration, \textsc{ARM}$\to$\textsc{DDLM}, consistently yields superior results compared to \textsc{DDLM}$\to$\textsc{ARM}. For instance, on the DART1 benchmark, LLada-8B~+~Llama-3.2-3B attains 73.0 versus 53.5, and on DART2, Dream-v0-7B~+~Llama3.1-8B achieves 61.0 versus 33.0.  
\textbf{(iii)} We hypothesize that this discrepancy arises from the planner’s (i.e., \textsc{DDLM}) tendency to generate non-fluent intermediate outputs, which may mislead the \textsc{ARM} executor. To validate this assumption and motivate our subsequent latent-space exploration, we perform a targeted diagnostic analysis described below.

\begin{table}[t]
\centering
\caption{Consolidated evaluation results on text-space collaboration across all model combinations and reasoning benchmarks. $\dagger$ Evaluated with \texttt{enable\_thinking=True},  $\ddagger$ with \texttt{enable\_thinking=False}.}
\label{tab:consolidated_performance}
\renewcommand{\arraystretch}{1.2}
\setlength{\tabcolsep}{1pt}
\rowcolors{3}{gray!5}{white}
\resizebox{\textwidth}{!}{%
\begin{tabular}{llccccccccc}
\toprule
\rowcolor{gray!10} \textbf{Model / Combination} & \textbf{Setup} & \textbf{ARC-E} & \textbf{ARC-C} & 
\textbf{DART-1} & \textbf{DART-2} & \textbf{DART-3} & \textbf{DART-4} & 
\textbf{DART-5} & \textbf{AIME} & \textbf{MMLU} \\

\midrule
\rowcolor{gray!20}\multicolumn{11}{c}{\textbf{ARMs only}} \\
\midrule

{Qwen2.5-3B} & ARM  & 91.0 & 86.5 & 58.0 & 34.0 & 29.5 & 17.5 & 11.5 & 1.0 & 61.0 \\
 & ARM $\to$ ARM  & 89.5 & 81.5 & 64.0 & 49.0 & 30.5 & 24.0 & 16.0 & 0.0 & 59.0 \\
\midrule

{Qwen2.5-7B} & ARM & 94.5 & 90.5 & 68.0 & 48.5 & 35.0 & 35.0 & 19.5 & 2.5 & 67.0 \\
 & ARM $\to$ ARM  & 96.5 & 89.0 & 72.5 & 56.5 & 35.0 & 33.5 & 20.0 & 3.0 & 61.0 \\
\midrule

{Llama-3.2-3B} 
 & ARM  & 87.5 & 79.5 & 44.5 & 35.5 & 28.0 & 34.5 & 36.5 & 3.5 & 54.0 \\
 & ARM $\to$ ARM  & 91.0 & 80.0 & 47.5 & 37.0 & 31.0 & 30.5 & 30.0 & 0.0 & 56.5 \\
\midrule

{Llama-3.1-8B} & ARM  & 87.5 & 79.5 & 68.5 & 50.5 & 36.0 & 36.0 & 20.5 & 2.5 & 63.5 \\
 & ARM $\to$ ARM  & 91.0 & 82.0 & 74.0 & 58.5 & 35.5 & 35.0 & 20.5 & 3.0 & 64.0 \\

\midrule

{DeepSeek-R1} & ARM  & 94.0 & 88.0 & 89.0 & 81.5 & 85.5 & 75.0 & 61.5 & 28.5 & 60.0 \\

\midrule

{Qwen3.1-7B}  & ARM$^{\dagger}$& 92.0 & 86.0 & 89.5 & 86.0 & 83.0 & 73.0 & 49.0 & 8.5 & 68.5 \\

\midrule
\rowcolor{gray!20}\multicolumn{11}{c}{\textbf{Dream-v0-7B setups}} \\
\midrule

\rowcolor{yellow!10} {Dream-v0-7B}  & DDLM  & \textbf{96.5} & 90.5 & \textbf{83.5} & 56.5 & 46.5 & \textbf{41.0} & 20.5 & 1.5 & 62.5 \\
  \rowcolor{yellow!10}& DDLM $\to$ DDLM& 95.0 & 91.5 & 76.5 & 59.0 & 44.0 & 38.5 & 19.0 & 3.0 & \textbf{65.5} \\
\midrule

{Dream-v0-7B + Qwen2.5-3B}  & ARM $\to$ DDLM & 92.0 & 87.0 & 79.0 & 59.0 & 47.0 & 34.5 & 17.0 & 2.0 & 61.5 \\
 & DDLM $\to$ ARM & 92.0 & 86.0 & 65.5 & 40.5 & 31.0 & 25.0 & 12.5 & 0.5 & 60.5 \\
\midrule

{Dream-v0-7B + Qwen2.5-7B} & ARM $\to$ DDLM & 96.0 & \textbf{92.0} & 76.0 & 61.0 & \textbf{48.0} & 40.0 & 17.5 & 2.5 & 63.5 \\
 & DDLM $\to$ ARM & 95.5 & 88.5 & 60.5 & 42.0 & 34.0 & 28.5 & 16.0 & 1.0 & 64.5 \\
\midrule

{Dream-v0-7B + Llama-3.2-3B} & ARM $\to$ DDLM & 95.0 & 89.5 & 74.5 & 60.0 & \textbf{48.0} & 40.5 & \textbf{23.5} & \textbf{3.0} & 60.5 \\
 & DDLM $\to$ ARM & 91.0 & 79.0 & 54.0 & 39.5 & 32.0 & 31.5 & 27.5 & 0.0 & 59.5 \\
\midrule

{Dream-v0-7B + Llama-3.1-8B} & ARM $\to$ DDLM & \textbf{96.5} & \textbf{92.0} & 77.0 & \textbf{61.0} & 47.5 & \textbf{41.0} & 16.0 & 2.0 & 62.5 \\
 & DDLM $\to$ ARM & 94.0 & 86.5 & 55.5 & 33.0 & 37.0 & 28.0 & 26.0 & 0.5 & 64.0 \\

\midrule
\rowcolor{gray!20}\multicolumn{11}{c}{\textbf{LLaDa-8B setups}} \\
\midrule

\rowcolor{yellow!10} {LLaDA-8B} & DDLM           & 94.5 & 87.5 & 80.0 & 59.0 & 41.0 & 35.5 & 15.0 & 1.5 & 59.0 \\
  \rowcolor{yellow!10}& DDLM $\to$ DDLM& 93.0 & 83.5 & 68.5 & 53.5 & 45.0 & 32.5 & 17.0 & 1.0 & 52.0 \\
\midrule

{LLaDA-8B + Qwen2.5-3B} & ARM $\to$ DDLM & 90.5 & 86.0 & 78.0 & 65.5 & 42.5 & 35.0 & 17.5 & 1.0 & 57.0 \\
 & DDLM $\to$ ARM & 88.5 & 85.0 & 68.0 & 44.5 & 29.0 & 23.0 & 14.0 & 0.5 & 55.5 \\
\midrule

{LLaDA-8B + Qwen2.5-7B} & ARM $\to$ DDLM & \textbf{95.0} & \textbf{91.0} & 80.5 & 61.5 & 48.0 & 31.5 & 18.5 & 1.0 & \textbf{61.5} \\
 & DDLM $\to$ ARM & 92.5 & 88.5 & 73.5 & 58.0 & 37.5 & 37.0 & 16.0 & \textbf{1.5} & 54.0 \\
\midrule

{LLaDA-8B + Llama-3.2-3B} & ARM $\to$ DDLM & 93.0 & 84.0 & 73.0 & 59.5 & 44.0 & 36.0 & 25.0 & 1.0 & 52.0 \\
 & DDLM $\to$ ARM & 90.5 & 82.5 & 53.5 & 43.0 & 35.5 & 30.0 & \textbf{27.0} & 0.0 & 52.5 \\
\midrule

{LLaDA-8B + Llama-3.1-8B} & ARM $\to$ DDLM & 91.5 & 86.5 & 81.0 & 62.0 & 48.0 & 32.0 & 20.0 & 1.0 & 60.5 \\
 & DDLM $\to$ ARM & 91.5 & 82.5 & 74.0 & 60.0 & 38.5 & 37.0 & 16.5 & \textbf{1.5} & 56.5 \\
\midrule

{LLaDA-8B + DeepSeek-R1} & DDLM $\to$ ARM & 92.5 & 88.5 & 73.5 & 58.0 & 37.5 & 37.0 & 16.0 & \textbf{1.5} & 54.0 \\
\midrule

{LLaDA-8B + Qwen3.1-7B} & DDLM $\to$ ARM$^{\ddagger}$ & 91.0 & 82.5 & \textbf{87.0} & \textbf{70.0} & \textbf{58.5} & \textbf{46.0} & \textbf{27.0} & 1.0 & 52.5 \\
\bottomrule

\end{tabular}
}
\end{table}



\begin{figure}[h]
    \centering
    \includegraphics[width=1.0\textwidth]{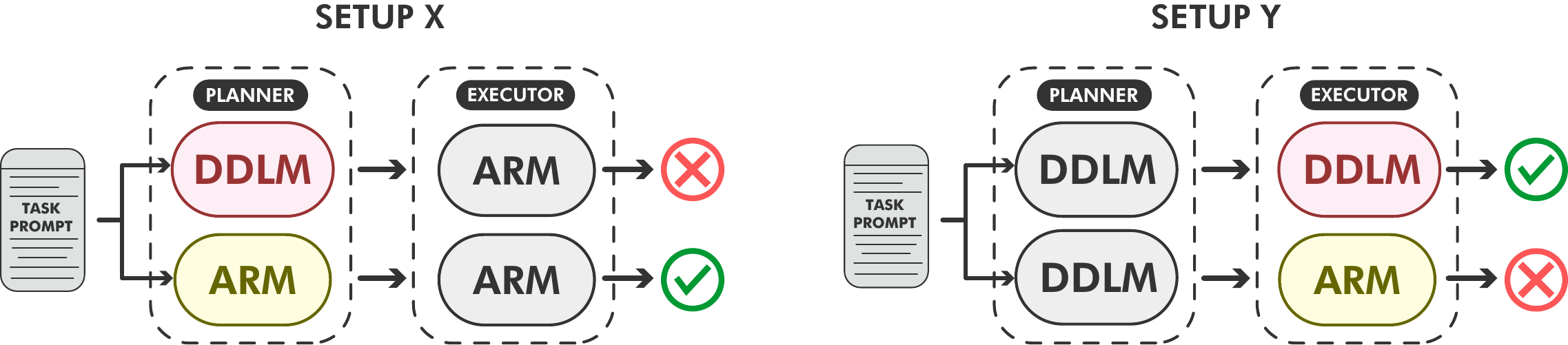}
    \caption{Diagnostic configurations for attributing errors to planner or executor. 
    \textbf{Setup X} tests whether failures stem from the planner: if replacing the diffusion planner (DDLM) with an autoregressive planner (ARM) fixes the output, 
    the error is attributed to the DDLM. 
    \textbf{Setup Y} tests executor reliability: if a diffusion executor succeeds where an ARM executor fails, the limitation lies in the executor.}
    \label{fig:setupxy}
\end{figure}

\subsection{Diagnosing Failure Modes}
\label{sec:diagnosing_failures}
To better interpret the outcomes of text-space collaboration and motivate the transition to latent-space collaboration, we conduct a diagnostic analysis of text-space outputs to assess whether the \(\text{DDLM} \to \text{ARM}\) configuration could benefit from latent-space intervention.
Specifically, we analyze two representative subsampled setups:  \textbf{(i) Setup X}: Questions where \(\text{DDLM} \to \text{ARM}\) fails due to planning errors (the DDLM), i.e., cases where the ARM could have succeeded given a more coherent plan. \textbf{(ii) Setup Y}: Questions where \(\text{DDLM} \to \text{DDLM}\) succeeds but \(\text{DDLM} \to \text{ARM}\) fails, indicating that the executor (ARM) is the limiting factor. This selection procedure is illustrated in Figure~\ref{fig:setupxy}.  
We then define diagnostic percentages to quantify planner versus executor contributions:
\vspace{-0.5em}
\[
\text{\textbf{Percentage}}_{i} = 
\frac{\# \{\text{Setup}\ {i} \text{ samples}\}}
{\# \{\text{Incorrect samples in } \text{DDLM} \to \text{ARM}\}} \quad \text{for } i \in \{X,Y\}.
\]

\begin{wrapfigure}{r}{0.53\textwidth}
  \centering
  \vspace{-1.0em} 
  \includegraphics[width=\linewidth]{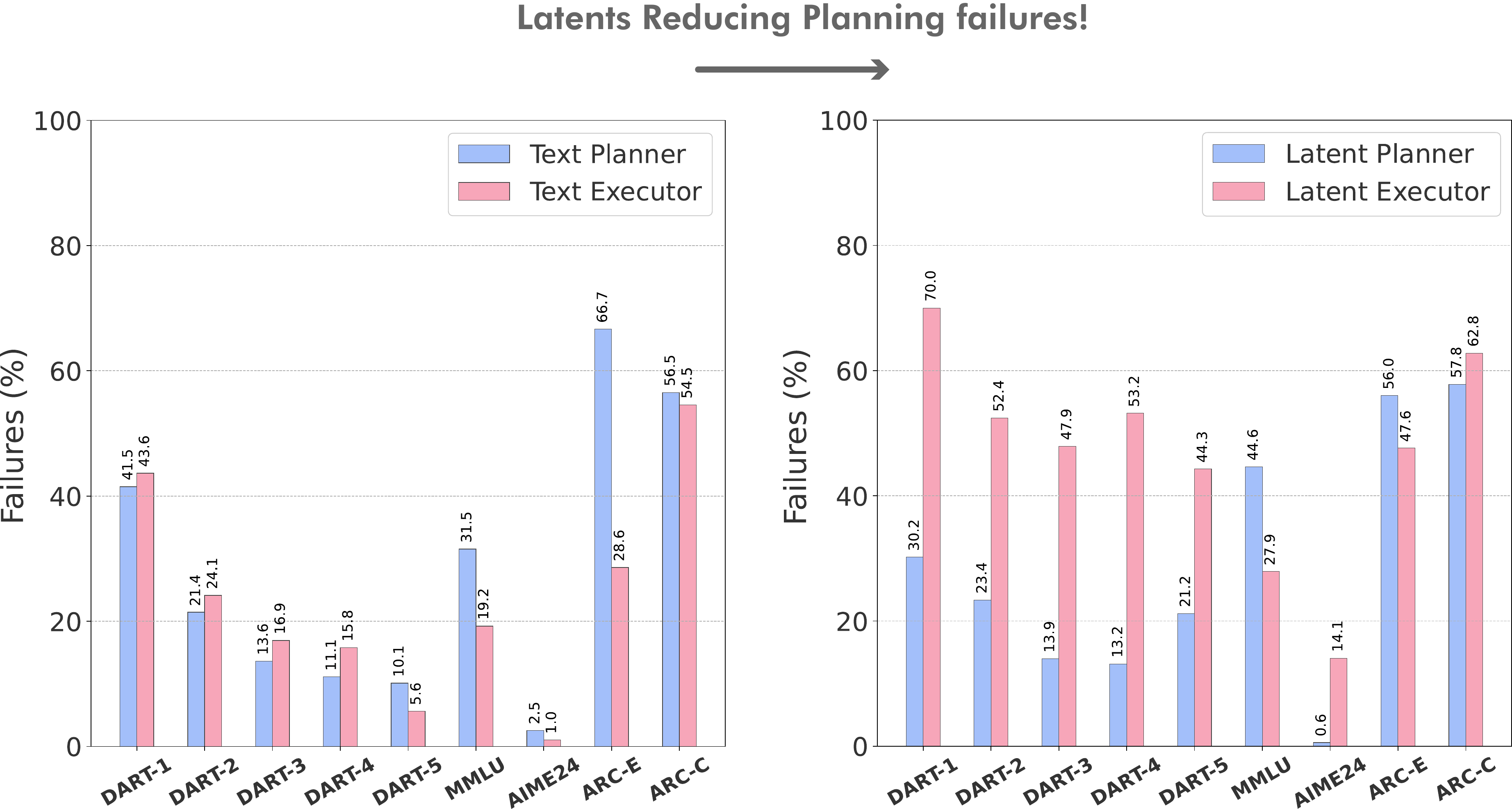}
  \caption{\textbf{Planner vs.\ executor failures in text- vs.\ latent-space collaboration.} 
  Results for LLaDA-8B-Instruct and Llama-3.1-3B-Instruct. 
  Latent-space collaboration substantially reduces planning errors compared to text-space.}
  \vspace{-0.8em}
  \label{fig:performance_comparison}
\end{wrapfigure}
These ratios measure the relative impact of planning versus execution errors in the collaboration pipeline. The performance comparison between Setup X (planner-focused) and Setup Y (executor-focused) across various benchmarks is presented in Figure~\ref{fig:performance_comparison}. The results reveal that, under text-space collaboration (DDLM$\rightarrow$ARM), 
the majority of percentage errors originate from the planner rather than the executor, 
indicating that the planner contributes more significantly to performance degradation. 
In contrast, under latent-space collaboration, after conducting the experiments described in Section~\ref{sec:latent_collaboration}, we perform the same diagnostic analysis and observe a notable shift: the executor emerges as the primary source of errors. This indicates that the gains achieved through latent-space collaboration predominantly enhance the planner component. The exact error distributions and percentages are reported in Appendix~\ref{appendix:comparison}.

\vspace{-0.8em}
\section{Collaboration Through Latent Space}
\label{sec:latent_collaboration}

Building on the diagnostic findings from text-space collaboration, we next investigate whether direct
information exchange in the latent domain can alleviate the limitations identified in the
\(\text{DDLM} \to \text{ARM}\) pipeline. The motivation is to bypass surface-level fluency constraints
of diffusion-generated text by allowing the planner and executor to communicate through aligned
latent representations. This section details the experimental configuration of our latent-space
collaboration framework and presents the resulting performance comparisons against text-based
counterparts.

\subsection{Experimental Setup}

To enable \(\text{DDLM} \to \text{ARM}\) collaboration, we train a Linear-GELU-Linear projector (Figure~\ref{fig:projector}) that maps DDLM hidden states (latents) into the ARM embedding space.  
We construct the training data by generating DDLM latent representations from 35K samples, drawn uniformly (5K each) from ARC\_Easy, ARC\_Challenge, DART1, DART2, DART3, DART4, and DART5.  For each sample, we produce latents with fixed output lengths of 64, 128 and 256 tokens.  
The projector is trained with both ARM and DDLM weights frozen. Training uses a cross-entropy objective, comparing ARM predictions, conditioned on latent-space inputs, with the corresponding ground-truth tokens from the benchmarks.
In our setup, we use LLaDa-8B-Instruct as the DDLM planner and Llama-3.2-3B-Instruct as the ARM executor.

\begin{table*}[t]
\centering
\caption{Evaluation of DeepSeek-R1-Distill-Qwen-7B and Qwen3-1.7B on reasoning benchmarks, including text-space vs. latent-space collaboration.}
\label{tab:model_performance}
\renewcommand{\arraystretch}{1.1}
\setlength{\tabcolsep}{5pt}
\resizebox{\textwidth}{!}{%
\begin{tabular}{lccccccccc}
\toprule
\rowcolor{gray!10}\textbf{Model / Setting} & \textbf{ARC-E} & \textbf{ARC-C} & \textbf{DART-1} & \textbf{DART-2} & 
\textbf{DART-3} & \textbf{DART-4} & \textbf{DART-5} & \textbf{AIME24} & \textbf{MMLU} \\
\midrule

\rowcolor{gray!20}\multicolumn{10}{c}{\textbf{Accuracy}} \\

DeepSeek-R1 (ARM only ) & 94.0 & 88.0 & 89.0 & 81.5 & 85.5 & 75.0 & 61.5 & 28.5 & 60.0 \\

Qwen3-1.7B (ARM only) & 92.0 & 86.0 & 89.5 & 86.0 & 83.0 & 73.0 & 49.0 & 8.5 & 68.5 \\

Llama-3.1-3B (ARM only) & 87.5 & 79.5 & 44.5 & 35.5 & 28.0 & 34.5 & 36.5 & 3.5 & 54.0 \\

LLaDA-8B $\to$ Llama-3.1-3B (text-space) & 90.5 & 82.5 & 53.5 & 43.0 & 35.5 & 30.0 & 27.0 & 0.0 & 52.5 \\

\midrule
\rowcolor{yellow!10} LLaDA-8B $\to$ Llama-3.1-3B (latent-space) &   &   &   &   &   &   &   &  &  \\
\rowcolor{yellow!10}\emph{64 tokens plan}            & 85.0 & 78.5 & 78.5 & 62.5 & 57.0 & 63.0 & 54.0 & 12.5 & 52.0 \\
\rowcolor{yellow!10}\emph{128 tokens plan}          & 87.5 & 76.5 & 70.5 & 43.0 & 43.0 & 49.0 & 36.0 & 14.0 & 44.0 \\
\rowcolor{yellow!10}\emph{256 tokens plan  }         & 85.0 & 81.0 & 70.0 & 62.0 & 50.0 & 62.0 & 52.5 & 14.0 & 15.5 \\

\midrule
\rowcolor{gray!20}\multicolumn{10}{c}{\textbf{Number of Tokens in average}} \\

DeepSeek-R1 (ARM only) & 398  & 504  & 1420 & 1833 & 2076 & 2418 & 3068 & 3832 & 760 \\

Qwen3-1.7B (ARM only) & 282  & 397  & 1024 & 1669 & 2112 & 2550 & 3106 & 4036 & 984 \\

Llama-3.1-3B (ARM only)  & 4 & 4 & 17 & 16 & 22 & 28 & 41 & 462 & 4 \\

LLaDA-8B $\to$ Llama-3.1-3B (text-space) & 4 & 4 & 10 & 14 & 16 & 12 & 20 & 503 & 4 \\

\midrule

\rowcolor{yellow!10}LLaDA-8B $\to$ Llama-3.1-3B (latent-space) &   &   &   &   &  &   &   &   &  \\

\rowcolor{yellow!10}\emph{plan (64, 128, 256 tokens) + executor :\textsuperscript{†} } & 2 & 2 & 4 & 5 & 5 & 5 & 6 & 14 & 2 \\

\bottomrule \multicolumn{10}{l}{\scriptsize\textsuperscript{†}\textit{Executor token averages; planner tokens (64, 128, 256) are to be added implicitly.}}
\end{tabular}
}
\end{table*}

\subsection{Results}

\begin{figure*}[h]
    \centering
    \includegraphics[width=0.95\textwidth]{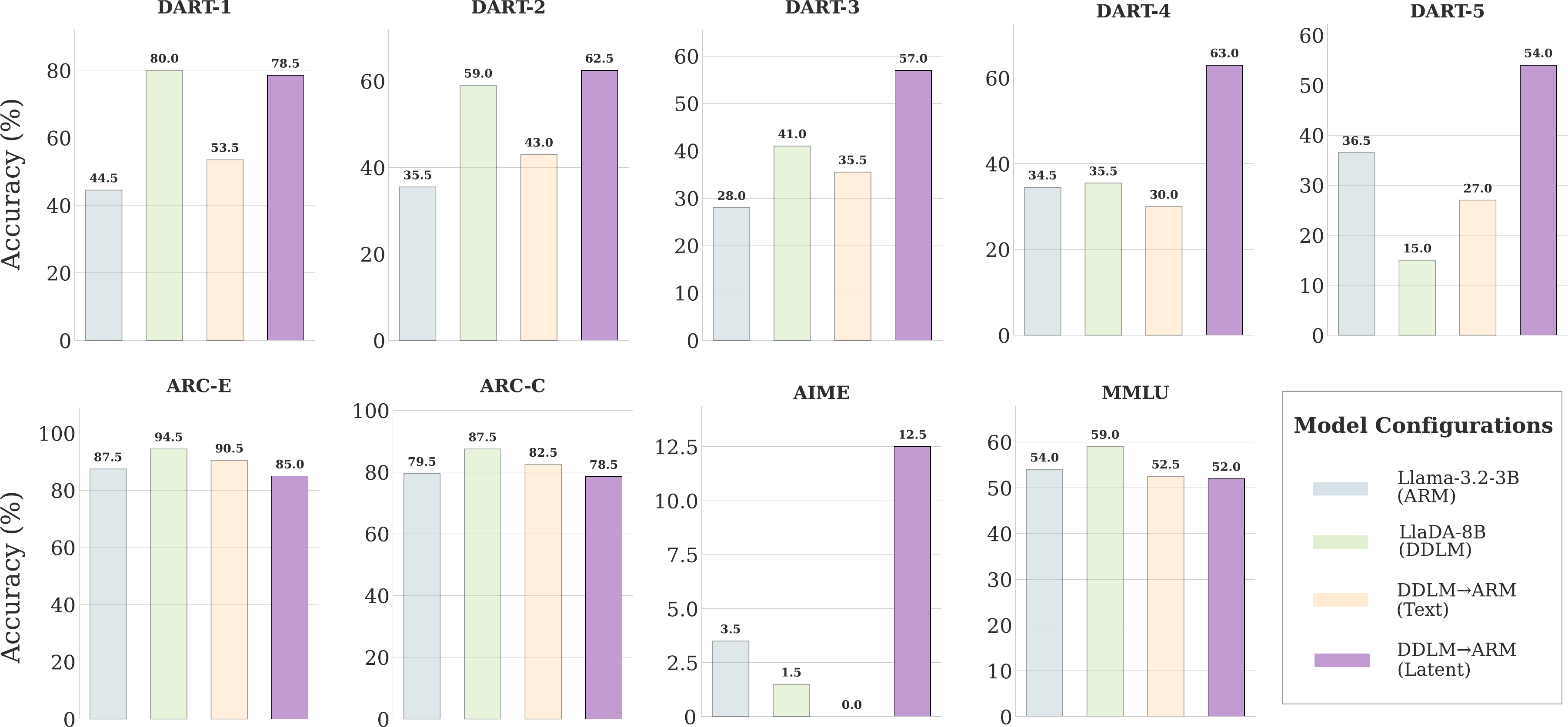}
    \caption{\textbf{Benchmark comparison of text-space vs.\ latent-space collaboration.} 
Accuracy of isolated models (LLaMA-3.2-3B ARM, LLaDA-8B DDLM) and collaborative configurations. 
In the latent setting, DDLM (64-token planner) combined with the ARM executor consistently outperforms text-space collaboration on DART and AIME, 
while maintaining comparable performance on ARC and MMLU.}

    \label{fig:latentsbar}
\end{figure*}

\paragraph{Latent-space vs Text-space Collaboration }
The results of the latent-space collaboration compared to the text-space baseline are reported in Table~\ref{tab:model_performance} and illustrated in Figure~\ref{fig:latentsbar}. While performance on ARC-E (85.0 vs.~90.5) and ARC-C (81.0 vs.~82.5) remains comparable across the two settings, the latent space consistently yields substantially higher accuracy on the DART benchmarks: e.g., DART-1 (78.5 vs.~53.5), DART-2 (62.5 vs.~43.0), DART-3 (57.0 vs.~35.5), DART-4 (63.0 vs.~30.0), and DART-5 (54.0 vs.~27.0). On AIME, the latent approach performance reaches 12.5\% with the projector trained on 64 tokens and 14\% with the projector trained on 128 or 256 tokens, compared to 0.0\% for the text space. Significantly, this improvement is obtained even though the projector between DDLM and ARM was trained without using any data from AIME or MMLU, yet it still generalizes to deliver markedly stronger results on these challenging evaluations. This highlights the promise of latent-space communication as an effective channel for planner–executor collaboration.

\paragraph{Token Budget and Efficiency}  
We explicitly control the length of diffusion-generated plans (64, 128, 256 tokens). Results demonstrate that longer is not always better:  Notably, the 64-token configuration offers the best overall trade-off between accuracy and efficiency on average. One plausible explanation for this superior performance in MMLU for instance, is its lower degree of redundancy. As shown in Appendix~\ref{sec:planner_repetition}, the repetition rate of diffusion outputs with 64 tokens in LLaDA-8B is comparable to that of Qwen2.5-7B and remains similarly low, whereas configurations with 128 and 256 tokens exhibit slightly higher repetition. 

Most importantly, latent space collaboration is markedly more efficient than both baselines 
(DDLM$\to$ARM in text space and ARM-only) as well as reasoning models. 
Remarkably, with only 64 planner tokens and an average of 5 executor tokens, it surpasses \textsc{Qwen3} on DART-5 while using merely \textbf{2.2\%} of the tokens, and outperforms it on AIME with just \textbf{1.9\%}. 
Although it does not yet reach the raw accuracy of \textsc{DeepSeek-R1}, it achieves highly competitive performance at a fraction of the computational cost, 
as reflected in the average token usage reported in Table~\ref{tab:model_performance}.

\section{Discussion}

Our investigation provides several insights into the collaboration between discrete diffusion language models (DDLMs) and autoregressive models (ARMs) under a planner–executor framework. 

First, the \textbf{communication channel} between planner and executor strongly shapes performance. Text-space collaboration is simple and interpretable, but constrained by the limitations of diffusion outputs, which ultimately prevents the collaboration from yielding measurable gains. Latent-space collaboration alleviates this bottleneck by allowing the executor to directly leverage the internal representations produced by the diffusion model.
This shift produces substantial accuracy gains on challenging benchmarks such as DART and AIME, where text-based prompting alone struggles. As shown by the diagnostic analysis in Subsection~\ref{sec:diagnosing_failures}, 
latent exchange reshapes the error landscape: while planner-driven failures dominate in text-space collaboration, 
executor-related errors become more prevalent in latent space, indicating that the gains primarily benefit the planner.

Second, \textbf{efficiency} emerges as a natural outcome of hybrid collaboration. Latent-space setups achieve competitive or superior accuracy to strong reasoning baselines while using one to two orders of magnitude fewer tokens. For example, with only 64 planner tokens plus $\sim$5 executor tokens, it surpasses Qwen3 on DART-5 and AIME, while using merely 2.2\% and 1.9\% of the tokens, respectively. This efficiency indicates that hybridization does not simply replicate autoregressive reasoning with fewer steps, but rather restructures the compute-accuracy trade-off in a qualitatively manner.

Finally, these results highlight several \textbf{open challenges}. While latent communication improves performance, it sacrifices interpretability, raising questions about how to inspect or align internal reasoning signals. Moreover, diffusion planners sometimes overproduce redundant steps, suggesting a need for adaptive mechanisms that balance global revision with conciseness. Future work may also explore scaling effects, alternate projection architectures, or training regimes that align planner and executor representations more tightly.

\section{Conclusion}
We presented a systematic study of planner–executor collaboration between discrete diffusion and autoregressive models for reasoning tasks. By comparing multiple setups and communication channels, we found that positioning DDLMs as planners and ARMs as executors, coupled with latent-space exchange, offers the most effective balance of accuracy and efficiency. This hybrid approach achieves substantial improvements on challenging mathematical benchmarks while drastically reducing token budgets, demonstrating a new path for budget-aware reasoning.

Beyond empirical gains, our findings highlight that \textit{hybrid reasoning is not merely a compromise between paradigms but a design opportunity}, a way to explicitly divide labor across compute efficiency, linguistic fluency, and reasoning robustness. 

Looking ahead, several avenues remain open. Improving the interpretability of latent communication, aligning the planner and executor through joint training objectives, and extending hybrid reasoning to multimodal and agentic contexts represent promising next steps. We view this work as a foundation for future systems that integrate diverse generation mechanisms into modular reasoning pipelines, enabling models that are both more efficient and more interpretable.

\bibliography{iclr2026_conference}
\bibliographystyle{iclr2026_conference}

\newpage
\appendix
\section*{Appendix}

\section{More Insights On Planner Repetition of tokens}
\label{sec:planner_repetition}

We will perform a qualitative assessment to identify prompt repetition errors in the planner text in the setup DDLM $\to$ ARM in the text space, using 256, 128 and 64 tokens for the LlaDa-8B-Instruct model, as well as Qwen2.5-7B-Instruct and Dream-v0-7B-Instruct for comparison. This analysis examines how increasing the number of planning tokens affects repetition, relative to an autoregressive model baseline.

We use the following metrics proposed in \citep{dragon}:
\begin{itemize}
    \item Distinct-3 (D-3)
    \item Repetition-4 (R-4)
    \item Lexical Repetition (LR-n)
    
\end{itemize}
\vspace{0.5cm}

\textbf{Distinct-3 (D-3)} calculates the percentage of unique 3-grams over all 3-grams. The value of Distinct-3 takes values between 0 and 1, with the closer to 1 indicating that the text is more diverse at the 3-gram level. Let \( D_3 \) be the number of unique 3-grams in the text and \( T_3 \) be the total number of 3-grams in the text. Distinct-3 is then computed by the following formula:

\[
\text{Distinct-3} = \frac{D_3}{T_3} \times 100
\]

\textbf{Repetition-4(R-4)} 
Let \( T \) be the total number of sentences in the text, \( R_t \) be the number of 4-grams repeated in a sentence \( t \), and \( I(x) \) be an indicator function (1 if \( x \) is true, 0 if \( x \) is false). Then Repetition-4 is calculated as follows:

\[
\text{Repetition-4} = \frac{1}{T} \sum_{t=1}^{T} I(R_t > 1) \times 100
\]

\textbf{Lexical Repetition (LR-n)} computes the average percentage of 4-grams that occur at least \( n \) times in the generated text. Let \( G \) be the total number of possible 4-grams in all texts and \( L_g \) be the number of repetitions of \( G \), then Lexical Repetition (LR-n) is calculated by the following formula:

\[
\text{Lexical Repetition} = \frac{1}{G} \sum_{g=1}^{G} I(L_g \geq n) \times 100
\]

\paragraph{Results}

\begin{table*}[ht]
\centering
\caption{Repetition Evaluation}
\small
\begin{tabular}{|l|c|c|c|}
\hline
\textbf{Simulation} & \textbf{D-3} & \textbf{R-4} & \textbf{LR-n} \\
\hline
  & $\uparrow$ & $\downarrow$ & $\downarrow$ \\
\hline
 \texttt{Qwen2.5-7B-Instruct (Baseline)} & 98.47   & 3.05  & 0.64 \\
\texttt{LLaDA-8B-Instruct (256 tokens)} & 83.05 & 10.52 & 6.74 \\
\texttt{LLaDA-8B-Instruct (128 tokens)} & 93.15  &  5.33 &  3.43 \\
\texttt{LLaDA-8B-Instruct (64 tokens)} &  96.85  & 3.37   & 1.33   \\
\texttt{Dream-v0-7B-Instruct (256 tokens)} &  62.02 & 3.26  & 2.15\\
\hline
\end{tabular}
\label{tab:fluency}
\end{table*}

In table \ref{tab:fluency}, the $\uparrow$ indicates that a larger value corresponds to better performance, while the $\downarrow$ indicates that a smaller value corresponds to better performance.

We observe a tendency for greater repetition in the plans generated by LLADA as the number of tokens increases.The number of repetitions in diffusion with 64 tokens in LLADA is comparable to that of Qwen2.5-7B and remains similarly low.

\section{Performance Comparison of Planner vs.\ Executor Issues}
\label{appendix:comparison}

\begin{table}[h]
\centering
\caption{Performance comparison of planner vs.\ executor issues for LLaDA-8B-Instruct and Llama-3.1-3B-Instruct under \textbf{Text-Space} vs.\ \textbf{Latent-Space} collaboration.}
\small
\renewcommand{\arraystretch}{1.2}
\begin{tabular}{l>{\columncolor{cyan!20}}c>{\columncolor{orange!20}}c>{\columncolor{gray!15}}c}
\toprule
\rowcolor{navy!80}
\textcolor{white}{\textbf{Benchmark}} & 
\textcolor{white}{\textbf{\begin{tabular}{@{}c@{}}Planning\\Failures (\%)\end{tabular}}} & 
\textcolor{white}{\textbf{\begin{tabular}{@{}c@{}}Execution\\Failures (\%)\end{tabular}}} & 
\textcolor{white}{\textbf{\begin{tabular}{@{}c@{}}Error Gap\\(\%)\end{tabular}}} \\
\midrule

\multicolumn{4}{l}{\textbf{\large \textcolor{blue!80!black}{LLaDA-8B + Llama-3.1-3B (Text-Space Pipeline)}}} \\
\midrule
DART-1        & 41.50 & \cellcolor{red!10}\textbf{43.64} & 2.14 \\
DART-2        & 21.43 & \cellcolor{red!10}\textbf{24.14} & 2.71 \\
DART-3        & 13.60 & \cellcolor{red!10}\textbf{16.92} & 3.32 \\
DART-4        & 11.11 & \cellcolor{red!10}\textbf{15.79} & 4.68 \\
DART-5        & \cellcolor{blue!10}\textbf{10.12} & 5.63 & 4.49 \\
MMLU          & \cellcolor{blue!10}\textbf{31.52} & 19.23 & 12.29 \\
AIME24        & \cellcolor{blue!10}\textbf{2.54}  & 1.03 & 1.51 \\
ARC-E         & \cellcolor{blue!10}\textbf{66.67} & 28.57 & 38.10 \\
ARC-C         & \cellcolor{blue!10}\textbf{56.52} & 54.55 & 1.97 \\
\midrule

\multicolumn{4}{l}{\textbf{\large \textcolor{purple!80!black}{LLaDA-8B + Llama-3.1-3B (Latent-Space Pipeline)}}} \\
\midrule
DART-1        & 30.23 & \cellcolor{red!10}\textbf{70.00} & 39.77 \\
DART-2        & 23.37 & \cellcolor{red!10}\textbf{52.42} & 29.05 \\
DART-3        & 13.95 & \cellcolor{red!10}\textbf{47.88} & 33.93 \\
DART-4        & 13.15 & \cellcolor{red!10}\textbf{53.19} & 40.04 \\
DART-5        & 21.21 & \cellcolor{red!10}\textbf{44.29} & 23.08 \\
MMLU          & \cellcolor{blue!10}\textbf{44.64} & 27.90 & 16.74 \\
AIME24        & 0.58  & \cellcolor{red!10}\textbf{14.07} & 13.49 \\
ARC-E         & \cellcolor{blue!10}\textbf{56.00} & 47.61 & 8.39 \\
ARC-C         & 57.80 & \cellcolor{red!10}\textbf{62.79} & 4.99 \\
\bottomrule
\end{tabular}
\end{table}

\section{Prompts for LLM Experiments}
\label{appendix:prompts}

\subsection{Planner Prompt}
\begin{verbatim}
You are a careful problem-solving planner.

Task: Produce ONLY a short list of HINTS that help solve the question. 
Do NOT state or imply the final answer. Do NOT mention any option letter 
(A, B, C, or D). Do NOT quote any option text verbatim. 
If you find yourself about to reveal a specific option or an answer, 
replace it with “[HIDDEN]”.

Format:
- Key facts to recall (2–4 bullets)
- Reasoning steps or elimination rules (2–5 bullets)
- Useful equations or definitions (if relevant)
- Edge cases or common traps (optional)

Be concise (<=120 words). No “Answer:” line. No letters A–D.

Question (stem only):
{question}
\end{verbatim}

\subsection{Executor Prompt}
\begin{verbatim}
You are an expert in solving multiple-choice questions.
Given the following plan or reasoning, please solve the question. 
If the plan contains any explicit answer or option letter, ignore it and 
solve from the hints + question only.

Plan:
{plan}
{question}
\end{verbatim}

\end{document}